\documentclass[letterpaper,twocolumn,fleqn]{article} 

\usepackage{ist}
\usepackage{graphicx}
\usepackage{amsmath,graphicx}
\usepackage{rotating,threeparttable,booktabs}
\title{A Psychological Study: Importance of Contrast and Luminance in Color to Grayscale Mapping}

\author{Prasoon Ambalathankandy $^{1}$, Yafei Ou $^{1}$, Sae Kaneko $^{2}$, Masayuki Ikebe $^{1*}$\\1 Research Center For Integrated Quantum Electronics, Hokkaido University, Japan\\2 Department of Psychology, Hokkaido University, Japan\\
Corresponding Author: ikebe@ist.hokudai.ac.jp$^{*}$}

\date{} % date has an empty field.

\begin{document} 

\maketitle 

\thispagestyle{empty} % prevents the first page to be numbered

%%%%%%%%%%%%%%%%%%%%%%%%%%%%%%%%%%
% Abstract
%%%%%%%%%%%%%%%%%%%%%%%%%%%%%%%%%%

\begin{abstract}
Grayscale images are essential in image processing and computer vision tasks. They effectively emphasize luminance and contrast, highlighting important visual features, while also being easily compatible with other algorithms. Moreover, their simplified representation makes them efficient for storage and transmission purposes. While preserving contrast is important for maintaining visual quality, other factors such as preserving information relevant to the specific application or task at hand may be more critical for achieving optimal performance. To evaluate and compare different decolorization algorithms, we designed a psychological experiment. During the experiment, participants were instructed to imagine color images in a hypothetical "colorless world" and select the grayscale image that best resembled their mental visualization. We conducted a comparison between two types of algorithms: 
(i) perceptual-based simple color space conversion algorithms, and (ii) spatial contrast-based algorithms, including iteration-based methods. Our experimental findings indicate that CIELAB exhibited superior performance on average, providing further evidence for the effectiveness of perception-based decolorization algorithms. On the other hand, the spatial contrast-based algorithms showed relatively poorer performance, possibly due to factors such as DC-offset and artificial contrast generation. However, these algorithms demonstrated shorter selection times. Notably, no single algorithm consistently outperformed the others across all test images. In this paper, we will delve into a comprehensive discussion on the significance of contrast and luminance in color-to-grayscale mapping based on our experimental results and analysis.
\end{abstract}
%
%\begin{keywords}
%	contrast, grayscale, luminance, naturalness, preprocessing
%\end{keywords}
%
\section{Introduction}
\label{sec:intro}
\begin{figure}[!t]
	\centering
	\includegraphics[width = 8.5cm]{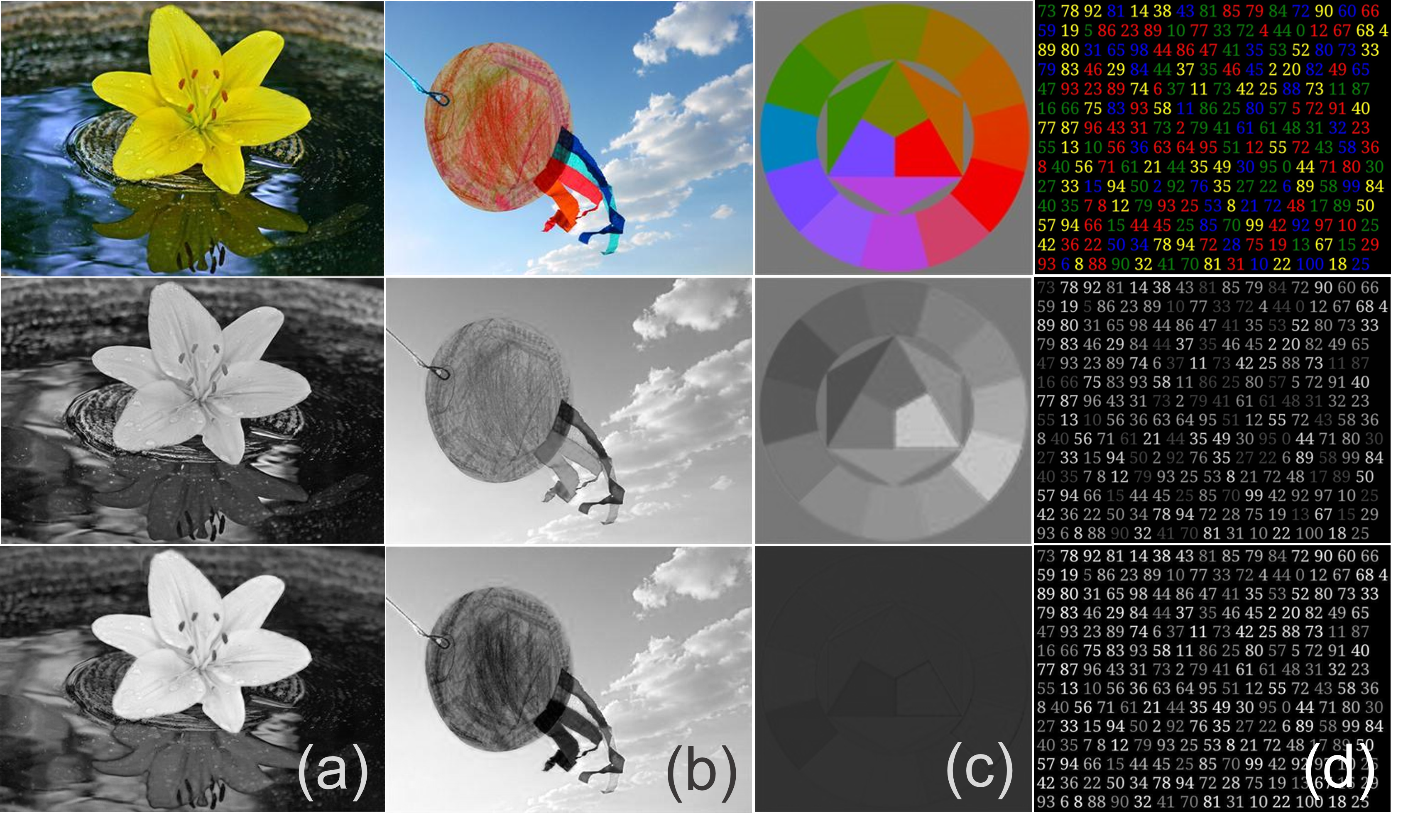}
\caption{Evaluation of Decolorization Algorithms: (Top Row) Original Color Images, (Middle Row) Grayscale Images Converted Using a 
	Perceptual Algorithm \cite{ambalathankandy2021warm}, (Bottom Row) (a) Kim et al. \cite{kim2009robust} - Demonstrating Loss of Chromatic Contrast. (b) Nafchi et al. \cite{nafchi2017corrc2g} - Highlighting Mishandling of Variations in Luminance. (c) CIELAB - Illustrating Failure with Isoluminant Colors. (d) Lu et al. \cite{lu2014contrast} - Displaying Inability to Handle Specific Color Combinations.}
	\label{fig_one}
\end{figure}

Grayscale images are widely used in various applications such as cost-efficient printing, medical displays, and e-ink devices. Many image processing algorithms, originally designed for color, are used with grayscale images to reduce memory and time complexity. However, this transformation leads to loss of information and as a result can reduce the range of the final resulting images as compared to the original. Decolorization methods compute grayscale images using luminance values from the source pixels \cite{akbulut2021new,liu2022two}. Figure \ref{fig_one} presents few decolorization examples from which we can learn their capabilities and limitations. The WCCD algorithm (warm-cool color-based decolorization) \cite{ambalathankandy2021warm} is showcased in the middle row of Fig. \ref{fig_one}, demonstrating its effectiveness in maintaining perceptual qualities. In contrast, (a) Kim et al. \cite{kim2009robust} exhibit a loss of chromatic contrast, (b) Nafchi et al. \cite{nafchi2017corrc2g} mishandle variations in luminance, (c) CIELAB fails with isoluminant colors, and (d) Lu et al. \cite{lu2014contrast} struggle with specific color combinations in the bottom row. Recent algorithms promise detailed grayscale output \cite{liu2022two}. However, they require more resources and time due to iterative processes thus, being unsuitable for practical real-time applications like video and printing. Decolorization algorithms aim to preserve color contrast in grayscale images, but this doesn't always optimize their performance in other image processing applications. Kanan and Cottrell studied the effect of grayscale images for image recognition application, and they conclude that the recognition performance is significantly affected by the underlying grayscales \cite{kanan2012color}. 

In this paper, we present the results of a comprehensive psychological experiment evaluating six distinct decolorization algorithms. Our evaluation includes three perception-based preprocessing methods (CIELAB, YCbCr, and WCCD \cite{ambalathankandy2021warm}) 
and three spatial contrast-based algorithms (\cite{lu2014contrast,nafchi2017corrc2g,liu2017log}). Our study underscores the pivotal roles played by contrast and luminance in the generation of grayscale images. Contrast and luminance preservation are crucial 
for maintaining visual quality when transitioning from color to grayscale. We chose not to include learning-based color-to-grayscale algorithms due to their black-box nature, which makes it challenging to understand the decision-making process behind color-to-gray mapping. Our findings reveal that perceptual algorithms excel in scenarios with limited color variance, where high color variance indicates a wide range of colors in the source image, whereas low color variance implies a more restricted color palette. Decolorization algorithms need to handle these variations appropriately to generate visually pleasing grayscale representations. Moreover, these perceptual methods offer simplicity, minimal parameter settings, and adaptability, making them valuable for video applications. This paper bridges the gap between perceptual quality and grayscale conversion techniques, enhancing our understanding of image processing applications.

\section{Related Works}
\label{sec:related}
Mapping 3D color information onto a 1D grayscale image while preserving the original appearance, contrast, and finest details is a non-trivial task, as depicted in Fig. \ref{fig_one}. There are several established methods for converting a color image into a grayscale one, and they can be categorized based on their mapping techniques (global and local methods), color space (RGB, LAB), computational complexity, and optimization methods, as reported by Akbulut et al. \cite{akbulut2021new}. While our primary focus in this work is on non-neural network-based methods due to concerns about their black-box nature, it's worth noting that there have been notable developments in learning-based color-to-gray algorithms. Convolutional Neural Networks (CNNs) have made notable strides in the field. Recent works include Cai et al.'s approach using a perceptual loss function \cite{cai2018perception}, Zhang et al.'s CNN framework for combining local and global features \cite{zhang2018contrast}, and Lin et al.'s use of partial differential equations (PDEs) for mapping color to gray \cite{lin2008learning}. However, they often face challenges related to control, computational complexity, and cost. These learning-based methods aim to generate high-quality grayscale images but come with their own set of limitations.

Luminance is fundamental to the human visual system, and several studies in the past have established that our visual acuity is dependent on changes in luminance \cite{graham1965vision}. Luminance also plays a significant role in perceiving a scene as natural \cite{jiang2017blind}. Choi et al. define naturalness as the level of resemblance between the photograph and the memories of real-life scenery \cite{choi2009investigation}. Additionally, Jiang et al. define naturalness as an image attribute based on the differences between normal exposure images and over-exposed or under-exposed images. To calculate their proposed tone mapping image quality metric, they compute the naturalness value based on luminance and yellow values \cite{jiang2017blind}. Luminance of an image can be measured from the pixel intensity, and in our experiment, we make use of the simple mean of the image as given by Eq. \ref{eq_lum}.
\begin{equation}\label{eq_lum}
	Luminance = \frac{1}{M \times N}\sum_{i=1}^{M}\sum_{i=1}^{N}(I_{i,j})
\end{equation}
Additionally, it's important to consider video decolorization, a related field with practical applications. Several video decolorization methods have been proposed in the past \cite{ancuti2011image, song2014real, tao2017video}. Tao et al. \cite{tao2017video} introduced three different decolorization strategies (low-proximity, median-proximity, and high-proximity) with varying computational costs and decolorization qualities. These methods address the unique challenges of converting video sequences into grayscale while preserving important visual information and offer insights into the complexities of real-time applications, which can be relevant to the field of image decolorization. This comprehensive overview of related works sets the stage for our discussion on contrast, luminance, and their significance in grayscale image generation.

\section{Psychological Experiment, Analysis and Results}
\label{sec:experiment} 

We conducted a thorough psychological experiment and analyzed the results. The experiment provided valuable insights into the performance and effectiveness of different decolorization algorithms. We believe the findings can guide readers in selecting appropriate algorithms for specific applications and contribute to the advancement of color to grayscale techniques.

\subsection{Setup, materials \& design}
\begin{figure}[!t]
	\centering
	\includegraphics[width = 8.5cm]{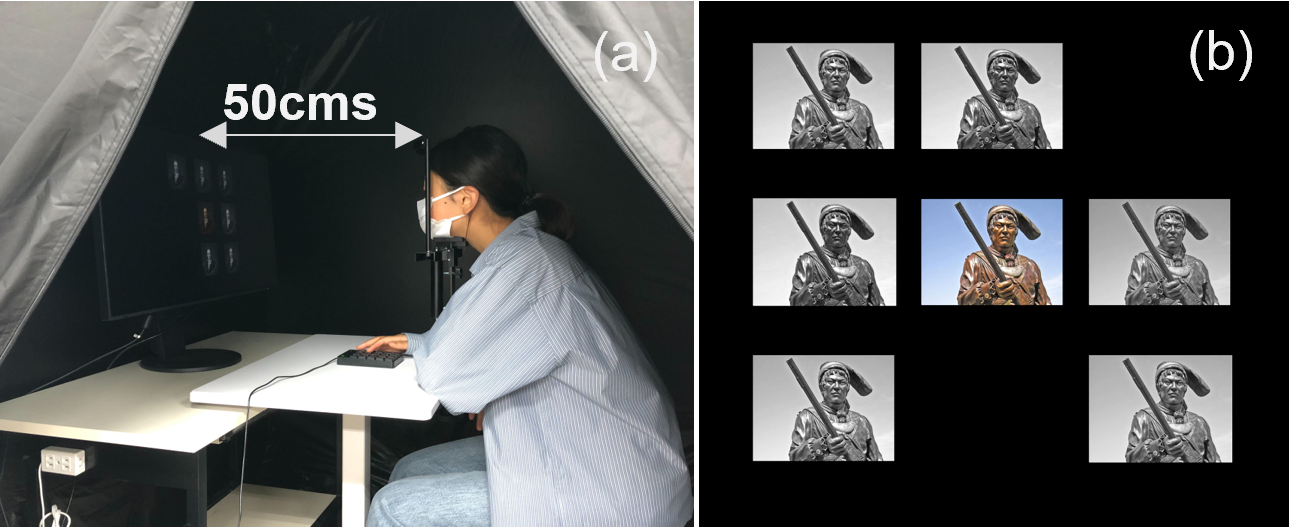}
	\caption{Our psychological experimental setup: (a) Participant preparing for the test. (b) Sample test display from our experimental dataset showing six grayscales surrounding the color image.}
	\label{fig_setup}
\end{figure}

We designed a MATLAB R2020b and Psychtoolbox (3.0.17)-based paired comparison graphical user interface (GUI) on an Ubuntu platform. GUI was used to display images and record participant's observations (image selection and timestamp). Our test images were loaded into computer memory and randomly generated pairs were displayed. The experiments were performed in a darkened room. The display monitor employed in our experiment was the EIZO EV3285 (3840 $\times$ 2160), and the viewing distance was fixed at 50 cm (viewing angle $\approx$ 34.92$^\circ$). The monitor was calibrated with a gamma setting of 2.2 and a white point calibrated to 6500K. Additionally, the brightness was set to 120. All the images used in our study were encoded in the RGB color space In order to avoid any artifacts due to head movements, we gently stabilized the participant’s head position by using a chin and forehead rest as shown in Fig. \ref{fig_setup}. The 53 test images for our experiments were selected from references \cite{lu2014contrast}, \cite{ cadik2008perceptual} and royalty free images from the internet. We recruited 54 naive observers (26 females, 28 males) aged between 22 to 50 who had normal or corrected-to-normal 20/20 vision from our university faculty, staff, graduate and undergraduate students’ community. The average time required to complete the test was about 21 minutes, and for their participation time all the participants were remunerated as per the university regulations. The only instruction we gave to the participants was to mentally visualize the presented color image in a ``\textit{colorless world}'' and pick one grayscale image from the six different grayscale images (CIELAB, YCbCr, WCCD \cite{ambalathankandy2021warm}, Lu et al.\cite{lu2014contrast}, Nafchi et al.\cite{nafchi2017corrc2g}, Liu et al.\cite{liu2017log} which resembles it the most.  

\subsection{Methodology and Hypothesis Testing}

Our experiment was conducted in accordance with our university's Code of Ethics regulations. The participants gave an informed and written consent to contribute to the experiment. A pretest examination was carried out to check for any color blindness among the participants using Ishihara test and one participant was excluded from the experiment due to color vision deficiency. 
During the psychological test when a participant chose a grayscale image it was scored 1, and the other 5 images were scored 0. Participants compared a color test image (at the center) with the six decolorized images (surrounding it), each of which were processed by a different algorithm. The participant response data were stored as a MATLAB data type $struct$ with fields recording participant number, observation table ($53\times12$) date and time of the experiment. The order, and location of grayscales were randomized to avoid any location-based bias and this information were stored in our observation table along with the user response time.  

In our paper, we have chosen algorithms for its distinct approach to color-to-grayscale conversion. CIELAB employs the CIELAB color space, focusing on perceptual uniformity and color-based conversion. YCbCr (MATLAB: rgb2ycbcr) utilizes a color space-centric approach and is commonly used in image and video processing. WCCD utilizes a customizable luminance calculation through adjustable bias parameters based on warm-cool colors \cite{ambalathankandy2021warm}. Lu et al.\cite{lu2014contrast} delves into contrast-preserving decolorization with an emphasis on global mapping methods. Nafchi et al.\cite{nafchi2017corrc2g} introduces a unique correlation-based approach, considering both color contrast and correlation. Lastly, Liu et al.\cite{liu2017log} adopts log-Euclidean metrics to preserve contrast during decolorization. These chosen methods collectively provide a comprehensive range of grayscale conversion techniques, facilitating a thorough evaluation across various applications, including perceptual quality assessment and video analysis.

We recorded 17,172 participant responses as 54 observers evaluated 53 images and their corresponding 6 grayscales. We analyzed the experimental data for all the color images individually and performed ${\chi}^2$ test(goodness of fit) \cite{sheskin2003handbook} 
to show that the participants choices were not random (i.e. all six grayscale images are equally likely to be chosen). The null hypothesis ($H_{0}$) of this test was that ``\textit{all grayscales were preferred equally}'', a scenario when the participants cannot tell the difference between grayscales, and/or, the images are all equally good. The alternative hypothesis ($H_{1}$) was ``\textit{all grayscales are not preferred equally}'', and the choice distribution was dependent on the perceptual quality of the grayscale image. For example, if the CIELAB grayscale transformation is consistently favored for a given test image, while the YCbCr transformation is never chosen, it provides a hint regarding the characteristics of an ``\textit{effective}'' color-to-grayscale conversion.

\begin{figure}[!t]
	\centering
	\includegraphics[width = 8.5cm]{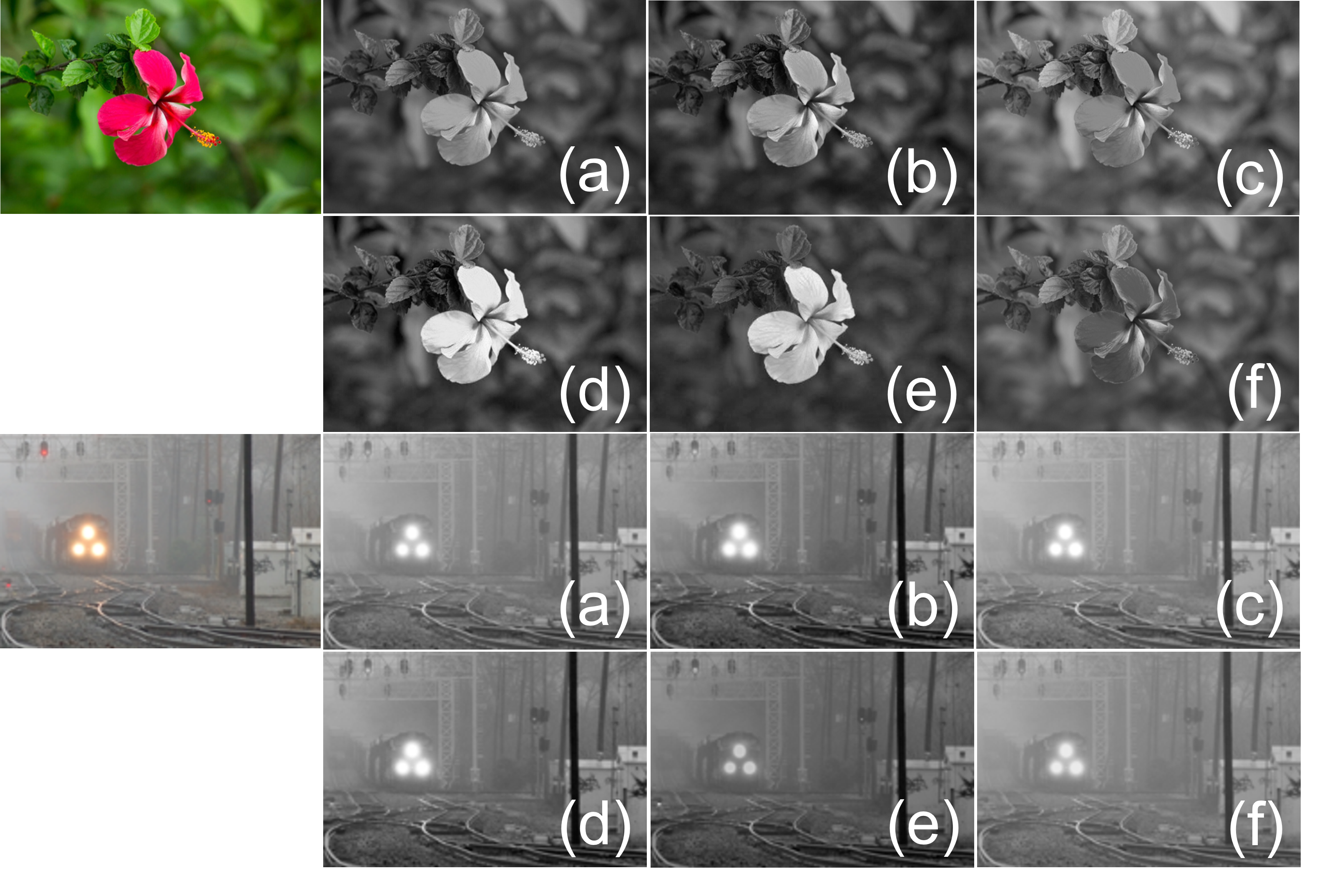}
	\caption{Grayscale Conversion Variability [(a) WCCD \cite{ambalathankandy2021warm} (b) Nafchi et al.\cite{nafchi2017corrc2g} (c) CIELAB (d) Liu et al.\cite{liu2017log} (e) Lu et al.\cite{lu2014contrast} (f) YCbCr]:
		Top Row - Typical case with a red flower image ${\chi}^2 = 26.77$. Bottom Row - Challenging case with a weather-degraded train image in foggy conditions, ${\chi}^2 = 2.33$, thereby resulting in acceptance of our null hypothesis ($H_{0}$).}
	\label{img44}
\end{figure}

In Fig. \ref{img44}, we present two test images and their grayscale conversions along with computed ${\chi}^2$ test statistics. 
The test statistic is computed using Eq. \ref{eq_chi}, where observed frequencies ($O_{i}$) represent user choices, and expected frequencies ($E_{i}$) indicate equal preference for all grayscale algorithms.
\begin{equation}\label{eq_chi}
	{\chi}^2=\sum_{i=1}^{k} \frac{(O_i - E_i)^2}{E_i}
\end{equation}
For instance, in the case of the red flower image (Fig. \ref{img44}a), we observed ${\chi}^2 = 26.77$, rejecting the null hypothesis. 
This indicates that grayscales were not equally preferred, with a significant preference for one or more algorithms. 
The degree of freedom ($df$) for this test is calculated as $5$ (i.e., $k-1$), where $k$ represents the number of grayscale algorithms.
To determine whether the ${\chi}^2$ values were statistically significant, we compared them with critical ${\chi}^2$ values for the 95th percentile (${\chi_{.05}}^2$) and 99th percentile (${\chi_{.01}}^2$) at $df = 5$. Since the calculated ${\chi}^2$ exceeded both ${\chi_{.05}}^2$ and ${\chi_{.01}}^2$, we rejected the null hypothesis at both $p<.05$ and $p<.01$ levels, indicating that participant choices were not random but influenced by grayscale quality. However, one image (Fig. \ref{img44}f) yielded a ${\chi}^2 = 2.33$, leading to the acceptance of our null hypothesis. We attribute this result to the image's complex scene content and a lack of clear cues, making it challenging for participants to distinguish grayscale quality effectively.

\begin{figure}[!t]
	\centering
	\includegraphics[width = 8.5cm]{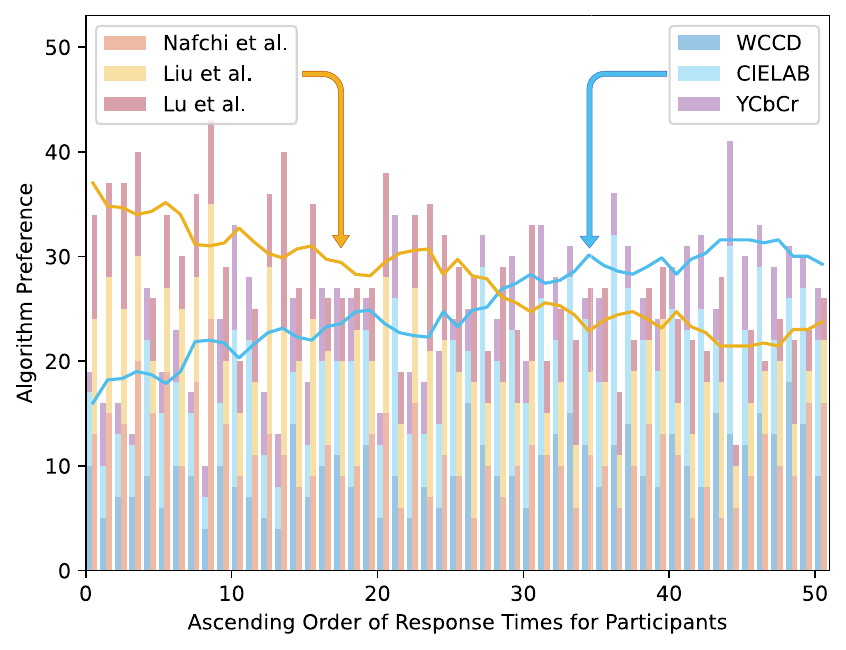}
	\caption{Participants with quick response times in seconds overwhelmingly preferred spatial contrast-based algorithms. User response times are arranged in ascending order, indicating the speed of decision-making.}
	\label{fig:contrast_Usertime}
\end{figure}

\subsection{Results \& discussion}
Several methods have been proposed to measure contrast, beginning with a simple ratio between maximum and minimum luminance of the scene, and choice of other metrics (like Weber contrast, Michelson contrast, and RMS contrast) depending on the end application. In this study we use RMS contrast which measures the standard deviation of pixel intensities and does not depend upon the spatial frequency content \cite{frazor2006local}. For an $M \times N$ image the RMS contrast is computed using Eq.\ref{eq_rms} where $I_{i,j}$ is the $i$-th, and $j$-th pixel intensity and $\bar{I}$ is the average intensity of the image. 
\begin{equation}\label{eq_rms}
	RMS_{contrast}= \sqrt{\frac{1}{M \times N}\sum_{i=1}^{M}\sum_{j=1}^{N}{(I_{i,j}-\bar{I})}^2}
	\end{equation}
We recorded the participants response (reaction) time, measuring the time it took for the observer to make a selection after the stimulus was presented. Upon arranging their response times in ascending order, we observed an interesting pattern, as shown in Fig. \ref{fig:contrast_Usertime}. We can clearly notice that the participants who made quick choices, preferred grayscale images generated by spatial contrast-based algorithms. We know that these algorithms employ techniques to enhance contrast, which can be lost during the dimensionality reduction process (color to gray). As a result, the grayscale images produced by these algorithms often exhibit high contrast, well-defined edges, and distinct textures, thereby improving visual salience. This relationship between contrast and salience has been studied by Cheng et al. \cite{cheng2014global}. The presence of good structures and edges in an image 
can improve the perceptual organization of visual objects, and this has been studied by Elder and Zucker \cite{elder1993effect}. Furthermore, studies by Itti et al. \cite{itti1998model} have reported that images with enhanced contrast, salient regions, and easily recognizable objects can lead to faster recognition. 

\begin{figure}[!t]
	\centering
	\includegraphics[width = 8.5cm]{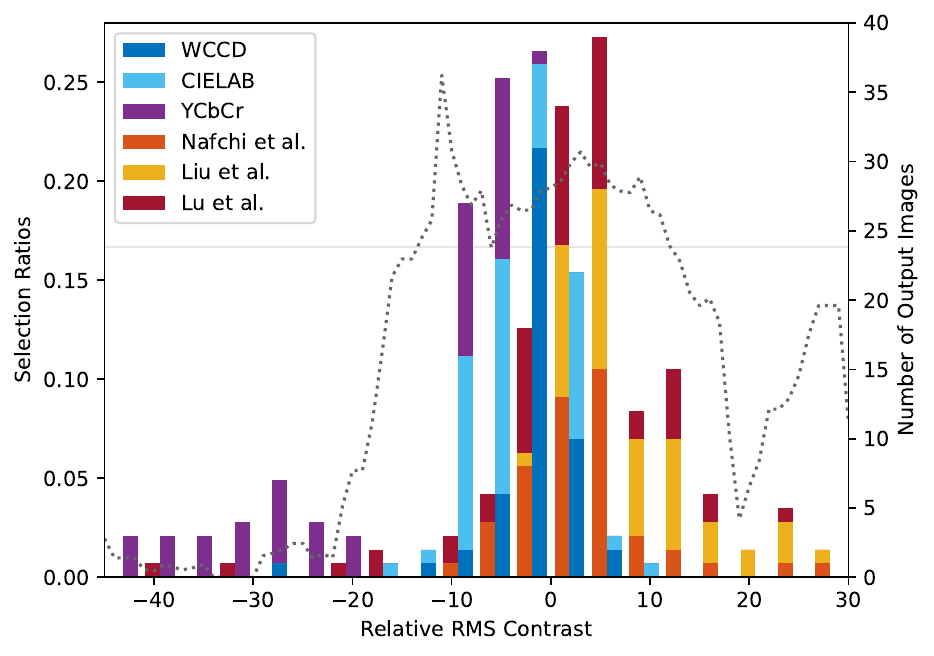}
	\caption{This visualization offers insights into user preferences and the distribution of output images based on relative RMS contrast. Spatial contrast-based methods enhance contrast in grayscale images. The left Y1-axis represents the curve plot, while the right Y2-axis represents the bar plot.}
	\label{fig:contrast_results}
\end{figure}

Our findings align with these studies and suggest that spatial contrast-based decolorization algorithms, emphasizing contrast enhancement, lead to quicker decision-making. However, this contrast-based approach compensates for this loss by introducing artificial contrast, as demonstrated in Fig. \ref{fig:contrast_results}. This figure illustrates the relative contrast difference between input and output concerning user selection. The selection ratio curve represents the average rate across all images under the current contrast level. Since we used six algorithms, the average selection ratio is $1/6$.

Nonetheless, spatial contrast-based algorithms may inadvertently emphasize details to the point of creating an exaggerated effect. In contrast, perceptual algorithms (CIELAB, WCCD, and YCbCr) avoid this by preserving the original contrast information. However, it's important to note that these claims are currently speculative. To investigate this hypothesis further, one potential approach is to filter the image using spatial bands, which would enable us to either confirm or refute this idea easily.

%The luminance is fundamental to human visual system, and several studies in the past have established that our visual acuity is dependent on the changes in luminance \cite{graham1965vision}. Luminance also plays a significant role to perceive a scene as natural \cite{jiang2017blind}. Choi et al. define naturalness as the level of resemblance between the photograph and the memories of the real-life scenery \cite{choi2009investigation}. And Jiang et al. define naturalness as an image attribute based on the differences of normal exposure images and over-exposed or under-exposed images. To calculate their proposed tone mapping image quality metric, they compute naturalness value based on the luminance and yellow values \cite{jiang2017blind}. Luminance of an image can be measured from the pixel intensity and in our experiment we make use of the simple mean of the image as given by Eq. \ref{eq_lum}.

%\begin{equation}\label{eq_lum}
%	Luminance = \frac{1}{M \times N}\sum_{i=1}^{M}\sum_{i=1}^{N}(I_{i,j})
%\end{equation}

\begin{figure}[!t]
	\centering
	\includegraphics[width = 8.5cm]{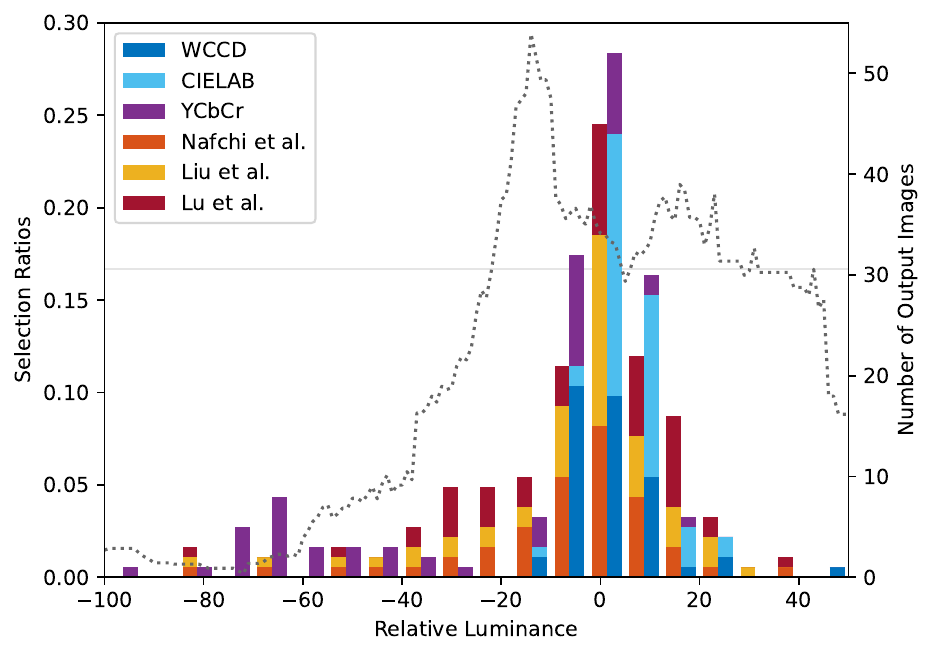}
\caption{This plot illustrates the impact of relative luminance on decolorization methods. Spatial contrast-based methods, as shown here, tends to neglect the significance of luminance and exhibit operations similar to DC-offset removal (see Fig. \ref{dcOffset_ver} for more details).}
	\label{fig:luminance_results}
\end{figure}

\begin{figure}[!t]
	\centering
	\includegraphics[width = 8.5cm]{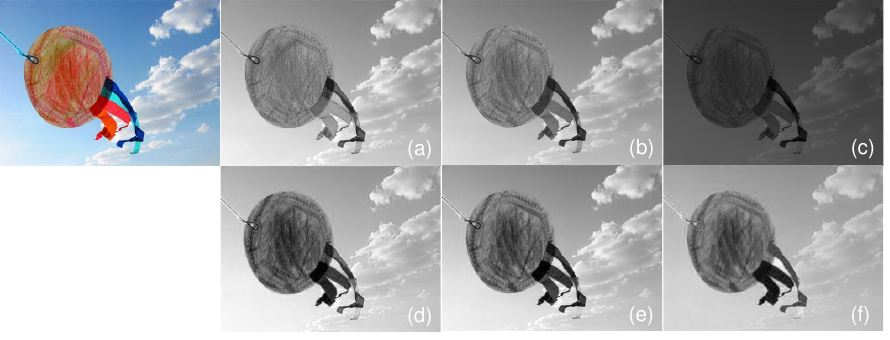}
\caption{Comparison of DC-offset removal in color-to-grayscale conversions using perceptually satisfactory simple color space algorithms (a, b, \& c) and spatial contrast boost algorithms (d, e, \& f). (a) WCCD \cite{ambalathankandy2021warm}, (b) CIELAB, (c) YCbCr, (d) Nafchi et al. \cite{nafchi2017corrc2g}, (e) Liu et al. \cite{liu2017log}, and (f) Lu et al. \cite{lu2014contrast}.}
	\label{dcOffset_ver}
\end{figure}

In Fig. \ref{fig:luminance_results} we plot the luminance difference between the input, output, and also include the reference to user selection rate. From this plot we can observe that the simple color space conversion algorithms retain the original luminance thereby invoking natural-like feeling, whereas the spatial contrast-based algorithms perform a DC-offset removal like operation. Mapping a light red kite tail with a sky blue background to darker shades of gray may be considered unnatural or undesirable in terms of preserving the visual appearance of the original scene as illustrated in Fig. \ref{dcOffset_ver}(d), (e), and (f). Grayscale images aim to represent the intensity or luminance information of the original colors, and mapping a distinct color combination to a significantly darker shade may result in a loss of information and visual fidelity. This suggests that the decolorization algorithm is not able to accurately capture the relative luminance values of the original colors, and it also indicates a limitation or deficiency in the algorithm's ability to handle specific color combinations or variations in luminance.

\subsection{Exploring color variance and video decolorization analysis}

In the following discussion, we compare the two decolorization methods: spatial contrast-boost algorithms and perceptual color space-based decolorization, 
with a specific focus on color variance and video decolorization. From our analysis, we find the unique characteristics and implications of each method in addressing these aspects. 

\begin{figure}[!t]
	\centering
	\includegraphics[width = 8.5cm]{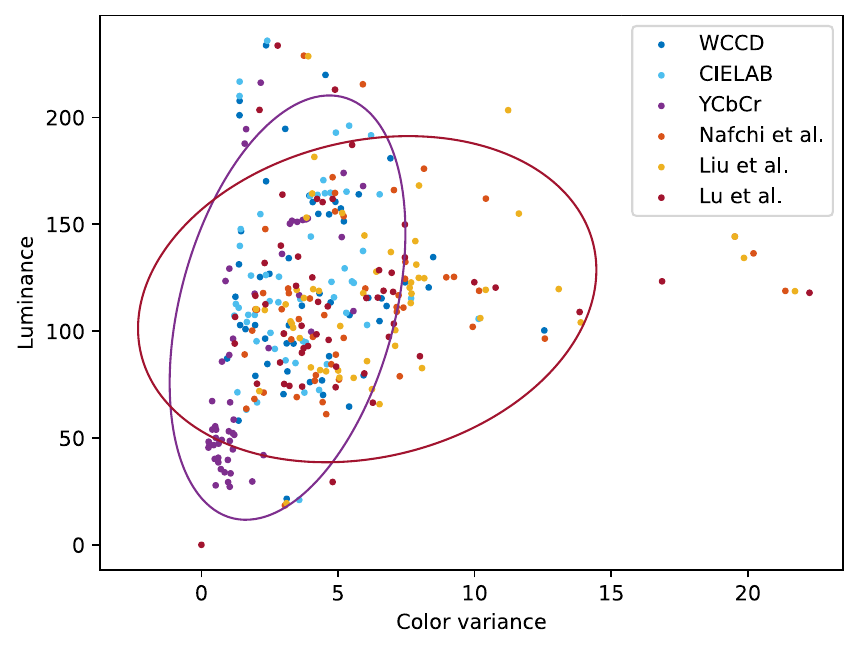}
	\caption{The scatter plot reveals that simple color space conversion algorithms exhibit better luminance preservation in grayscale images with low color variance, compared to spatial contrast-boost algorithms. This finding underscores the importance of considering the color variance in selecting the appropriate decolorization algorithm for optimal luminance representation.}
	\label{fig:var_lumi}
\end{figure}

\textit{\textbf{Measuring Color Variance:}} To assess color variance, we compute the standard deviation of color values within an image. This metric evaluates color variation across the image, assuming that changes in color are minimally influenced by shifts in luminance. Our approach involves converting RGB images to the YCbCr color space, which segregates luminance (Y) from chrominance (Cb and Cr). This separation proves especially practical for video applications. In Fig. \ref{fig:var_lumi} we plot the color variance versus luminance and observe that the simple color space conversion algorithms are practical candidates to generate grayscales for low variance images with good luminance information.  

\begin{figure}[!t]
	\centering
	\includegraphics[width = 8.5cm]{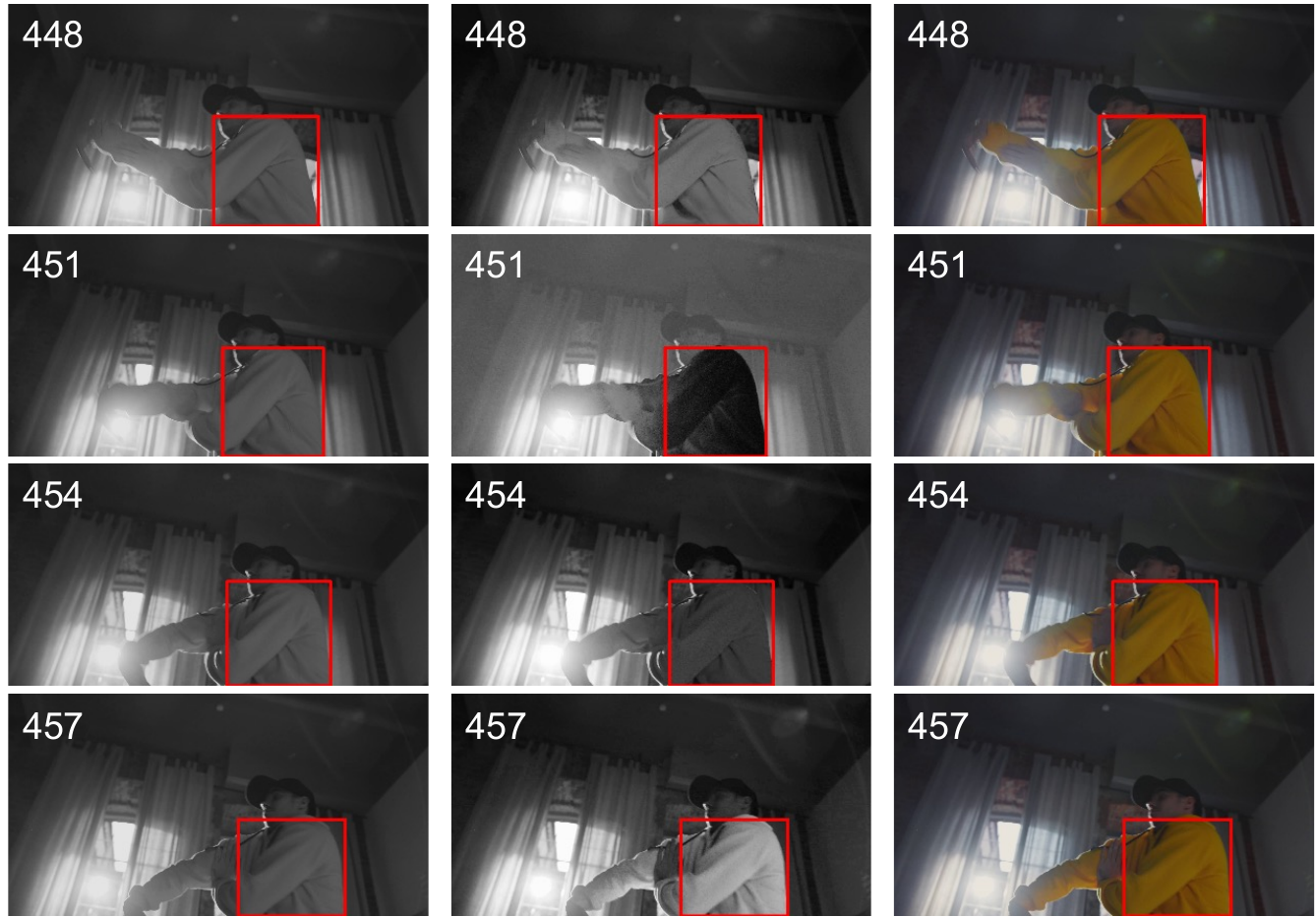}
	\caption{Comparison of video frames (original video courtesy of \cite{video_tran}): (left)  Grayscale frames using the perceptual-based algorithm (WCCD). (middle) Grayscale frames using the spatial contrast-based algorithm \cite{nafchi2017corrc2g}. (right) Original video.}
	\label{fig:vidTran}
\end{figure}

\begin{figure}[!t]
	\centering
	\includegraphics[width = 8.5cm]{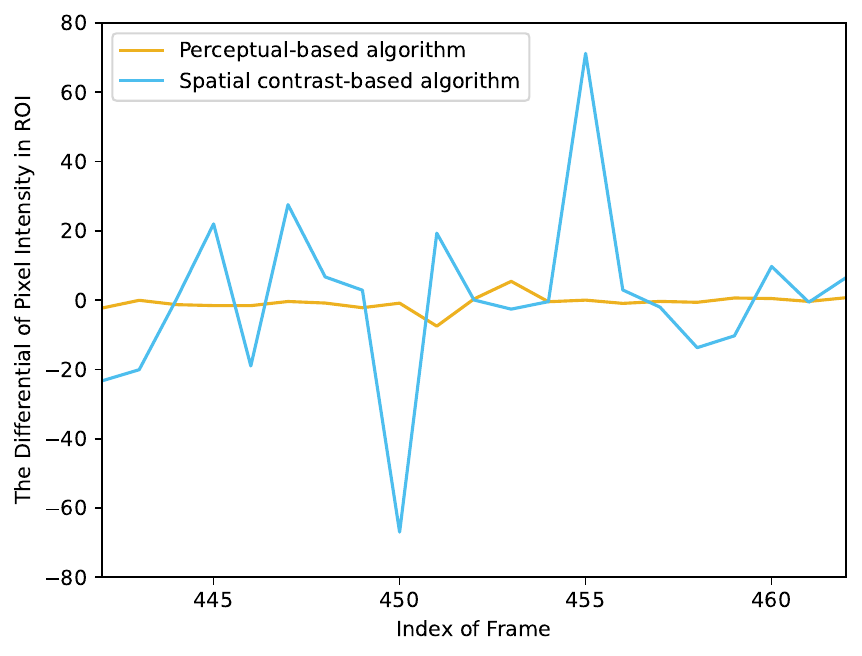}
	\caption{The 1D plot illustrates the intensity values of the region of interest (ROI) in the video shown in Fig. \ref{fig:vidTran}, obtained using two different decolorization algorithms. The spatial contrast boost decolorization algorithm \cite{lu2014contrast} exhibits flicker artifacts, resulting in fluctuations in intensity values. In contrast, the perceptual decolorization algorithm (WCCD) demonstrates a smoother transition of mean intensities, indicating its ability to preserve a visually pleasing and consistent intensity representation.}
	\label{fig:vid_roi}
\end{figure}

Perceptual decolorization algorithms, such as CIELAB, WCCD, or YCbCr, offer several benefits and advantages for video decolorization: (i) These algorithms are designed to preserve the luminance information from the color image, maintaining the brightness variations and luminance characteristics of the original scene. (ii) Additionally, they take into account the color perception characteristics of the human visual system. By considering factors like color spaces that align with human perception, these algorithm mappings ensure a consistent and accurate representation of colors in grayscale. This result in generating visually pleasing, natural-feeling, and coherent videos. Furthermore, the characteristics like  simplicity, low parameter dependency, and compatibility, make them well-suited for video applications.

On the other hand, spatial contrast-based algorithms prioritize local contrast, edge enhancement, and emphasize objects and details based on their spatial characteristics. These algorithms typically apply a uniform contrast enhancement across the entire image or regions of interest. However, this approach is ineffective in adapting to variations in image content, resulting in undesirable grayscale mappings as demonstrated in Fig. \ref{fig:vidTran}. Furthermore, this operation may potentially lead to abrupt changes in contrast and introduce flicker artifacts due to localized contrast enhancements resulting in pixel intensity variations, as shown in Fig. \ref{fig:vid_roi}. However, the actual performance and presence of flickering can vary depending on the specific algorithms, parameters, and characteristics of the input video.

%gray_ROI_1

\section{Conclusion}
\label{sec:conc}
Grayscale images are essential for many image processing applications. Despite the existence of numerous proposed solutions, determining the most suitable one for a specific application remains a challenge, and it often requires a subjective evaluation. In order to address this challenge, we conducted a psychological experiment to compare and evaluate two types of algorithms: (i) simple color space conversion algorithms and (ii) spatial contrast-based algorithms. The results of our experiment demonstrate that, on average, CIELAB performs better, and it highlights the need for preserving luminance information for achieving a natural appearance. Interestingly, we observed that individuals with quick response times tend to prefer spatial contrast-based algorithms. However, it is important to note that spatial contrast-boosting algorithms may result in an overemphasis of grayscales, which can be perceived as exaggerated. By conducting this psychological experiment and analyzing the data, our findings highlights the significance of preserving luminance information and maintaining appropriate contrast levels in grayscale images to ensure the visual quality. 

\section{Acknowledgment}
We sincerely thank the authors of \cite{lu2014contrast, nafchi2017corrc2g, liu2017log, kim2009robust, cadik2008perceptual} for making their datasets and source code available. We would also like to express our gratitude to the creators of the video \cite{video_tran}.

\bibliographystyle{IEEEbib2}
\bibliography{refs}

\end{document}